\documentclass[a4paper, 9pt]{article} 

\usepackage[dvipdfmx]{graphicx, color}
\usepackage{here}
\usepackage{amsmath}
\usepackage{amssymb}
\usepackage{booktabs} 
 \usepackage{ascmac}
\usepackage{geometry}
\usepackage{stmaryrd}
\usepackage {cancel}
\usepackage{setspace}
\usepackage{theorem}
\usepackage{url}
\usepackage{mathtools}

\newtheorem{defi}{Definition}[section]
\newtheorem{lemm}{Lemma}[section]
\newtheorem{chr}{Property}[section]

\newcommand{\argmin}{\mathop{\rm arg~min}\limits}

\usepackage{newtxtext,newtxmath}

\begin{document}

\singlespacing
\setstretch{1} 

\title{\vspace{-0cm}Alternative Methods to SHAP Derived from Properties of Kernels: A Note on Theoretical Analysis 
} 
\begin{spacing}{1.2}
\author{
Kazuhiro Hiraki
\thanks{kazuhiro.hiraki86@gmail.com}  \hspace{1cm}
Shinichi Ishihara 
\thanks{ishihara5683@gmail.com}\hspace{1cm}
Junnosuke Shino
\thanks{\textbf{Corresponding author}: Waseda University, 
junnosuke.shino@waseda.jp} 
}
\date{}
\maketitle
\end{spacing}
\begin{abstract}
\begin{spacing}{1}
This study first derives a general and analytical expression of AFA (Additive Feature Attribution) in terms of the kernel in LIME (Local Interpretable Model-agnostic Explanations). 
Then, we propose some new AFAs that have appropriate properties of kernels or that coincide with the LS prenucleolus in cooperative game theory. We also revisit existing AFAs such as SHAP (SHapley Additive exPlanations) and re-examine the properties of their kernels. 
\end{spacing}
\end{abstract}

\section{Introduction}
In the field of machine learning, Explainable Artificial Intelligence (XAI) refers to techniques and methods that make the decisions and predictions of machine learning models easier to understand. Among them, AFA (Additive Feature Attribution) is a method that decomposes a model's prediction into the contributions of individual features. Notably, SHAP (SHapley Additive exPlanations), proposed by \cite{lundberg2017}, which is based on the Shapley value \cite{shapley} in cooperative game theory, is well-known in this context. Recently, research on SHAP has been rapidly expanding (\cite{survey}). To reduce the computational cost of SHAP, various methods such as Tree-SHAP\cite{lundberg2017} and Fast SHAP \cite{fastshap} have been proposed and applied to actual data (for example, \cite{goldprice}). As an alternative to SHAP, \cite{Condevaux2023} considers ES (Equal Surplus) and FESP (Fair Efficient Symmetric Perturbation), both of which are based on solution concepts in cooperative game theory.

In this study, we investigate the relationship between AFA and the kernel in LIME (Local Interpretable Model-agnostic Explanations) as proposed by \cite{lime}.
\cite{lundberg2017} characterizes SHAP in terms of the kernel (Kernel SHAP) and derive the expression of SHAP kernel explicitly. Intriguingly, the properties of the SHAP kernel seem different from those that the LIME kernel is expected to have. More specifically, in LIME, the kernel attaches a large weight as a perturbed sample gets closer to the instance being explained, which is different for that of SHAP. In this note, we first provide a general framework to relate an AFA with its associated kernel by deriving an analytical expression of an AFA in terms of its kernel. Then, we propose some new AFAs that have reasonable properties of kernels or that coincide with the LS prenucleolus in cooperative game theory. We also revisit existing AFAs such as SHAP and reexamine the properties of their kernels.

\section{Preliminaries}\label{Prep}
Let $\textit{t}$ and $n$ be the number of the instances and the number of features, respectively. Suppose $N=\{1,...,n\}$, $T=\{1,...,t\}$. The feature input is a $t \times n$ matrix $X= (X_1,...X_j,...,X_n)$. The $j$th feature vector is $X_{j}=(x_{1,j},...,x_{t,j})'$ and, for the $\tau$th instance of interest, the vector of features is $x_{\tau} = (x_{\tau, 1},...,x_{\tau, j},..., x_{\tau, n})$. Let $f$ be the original prediction model which takes $x_{\tau}$ and produces a prediction. Let $Y=(y_1,..., y_t)'$ be the vector of the predicted values ($Y=f(X)$).

For an element of the power set of $N$, which is called a coalition in the cooperative game theory, $S\in 2^{N}$, define $x_{\tau, S} = \{x_{\tau, j} | j\in S\}$. $x_{\tau, S}$ is a vector that consists of features in $S$ at $\tau$th instance. Similarly, for $S\in 2^{N}$, define $X_S=\{X_j | j\in S\}$. 

In cooperative game theory, a characteristic function form game is expressed as $(N, v)$ where $N=\{1,...,n\}$ is the set of players and $v$ is a real-valued function on the power set $2^N$. For the $\tau$th instance and any coalition $S\in 2^N$, when we define $v_{\tau}: 2^N\longrightarrow \mathbb{R}$ as in (\ref{vtau}), a characteristic function form game $(N, v_{\tau})$ is specified for $\tau$:

\begin{equation}\label{vtau}
v_{\tau}(S) =E\left[f(x_{\tau, S}, X_{N\setminus S}) \right].
\end{equation}

$v_{\tau}(S)$ is interpreted as the prediction that $f$ produces for the $\tau$th instance, when (i) features $x_{\tau, j}$ where $j\in S$ are known but (ii) features $x_{\tau, k}$ where $k\in N\setminus S$ are unknown. Note that $v_{\tau}(N)= E\left[f(x_{\tau, 1},..., x_{\tau, n}) \right] = f(x_{\tau, 1},..., x_{\tau, n})$ and $v_{\tau}(\emptyset)= E\left[f(X_{1},..., X_{n}) \right] = E\left[f(X) \right]$, where the former is the prediction when all features at $\tau$th instances are known and the latter is the prediction when none of the features are known. It should be noted that, while standard cooperative game theory assumes that $v(\emptyset)=0$, this is not necessarily satisfied under this machine learning (ML) setting.

With this setup, Additive Feature attribution (AFA) is the method to decompose $v_{\tau}(N) - v_{\tau}(\emptyset)$ into features at $\tau$, depending on their \lq\lq contributions.\rq\rq \  More precisely, for a characteristic function form game $(N, v_{\tau})$ associated with the $\tau$th instance and for the feature (player) $j$, define a real-valued function $\Psi_{\tau} (j): N\longrightarrow \mathbb{R}$. We hereafter use $\Psi_{\tau} (j)$ and $\Psi_{\tau, j}$ interchangeably and let $\Psi_{\tau} = (\Psi_{\tau,1},...,\Psi_{\tau,n})$. When $\Psi_{\tau}$ satisfies $\sum_{j\in N}\Psi_{\tau, j}=v_{\tau}(N) - v_{\tau}(\emptyset)$, then $\Psi_{\tau}$ is called Additive Feature Attribution (AFA), denoted by $\Psi_{\tau}^{AFA}$. 

\section{A brief review on LIME and kernel}
Here we review \cite{lundberg2017} and \cite{lime}, specifically the parts concerning the relationship between LIME and SHAP. In their notation, $x$ is the original representation of an instance being explained and $z$ is a perturbed sample from $x$. They use a binary vector $x'$ and a mapping $x=h_x(x')$, but in this study, just for simplicity, $x=x'$ and $z=z'$ i.e., the original instances are \textit{simplified} (\cite{lundberg2017}), \textit{interpretable} (\cite{lime}) or binary from the beginning.   

\cite{lundberg2017} considers the following minimization problem (LIME, proposed by \cite{lime}). 
\begin{equation*}\label{minimization}
\xi(x) = \argmin_{g\in G} L(f, g, \pi_{x})+\Omega(g) \ \ \ \ \ where 
\end{equation*}
\begin{itemize}
\item $f$: the original prediction model.
\item $g$: the explanation model defined as $g(z)=\phi_0+\sum_{i=1}^{n}\phi_i z_i$, where $\phi_i\in \mathbb{R}$ and $n$ is the number of the features. Let $G$ be the set of all $g$s
and let $\phi =(\phi_1,...,\phi_n)\in \mathbb{R}^n$.
\item $x$: instance being explained.
\item $z$: perturbed sample from $x$. Let $Z$ be the set of all $z$s, including $x$.
\item $\pi_{x}$: local kernel. 
\end{itemize}
and $L$ is the loss function of the minimization problem and $\Omega(g)$ is a measure of complexity of $g$ (a more complex $g$ is penalized). Note that, regarding the kernel $\pi_{x}$,  \cite{lime} assumes it increases as the distance between $x$ and $z$ decreases, that is, as $z$ gets closer to $x$, a larger weight is attached to $z$.

Based on this setup, \cite{lundberg2017} assumes $\Omega(g)=0$ and $L(f, g, \pi_x)=\sum_{z \in Z} \left[f(z)-g(z)\right]^2\pi_x(z)$. Therefore, the minimization problem of (\ref{minimization}) is:

\begin{eqnarray}\label{minimization}
\argmin_{g\in G} \sum_{z \in Z} \left[f(z)-g(z)\right]^2\pi_x(z) & = & \argmin_{\phi \in \mathbb{R}^n} \sum_{z \in Z} \left[f(z)-\left\{\phi_0+\sum_{i=1}^{n}\phi_i z_i\right\}\right]^2\pi_x(z) \notag \\
\ & = & \argmin_{\phi \in \mathbb{R}^n} \sum_{z \in Z} \left[\sum_{i=1}^{n}\phi_i z_i-\left\{f(z)-\phi_0\right\}\right]^2\pi_x(z). \label{tochu}
\end{eqnarray}

Now recall $z$ is perturbed sample from $x$ and $x\in Z$. Therefore, summation over $Z$ in (\ref{tochu}) coincides with the summation over $2^N$ under our notation, and the summation of $\sum_{i=1}^{n}\phi_i z_i$ coincides with $\sum_{i\in S}\phi_i$. Therefore, under our notation, (\ref{tochu}) falls into the following: 

\begin{equation}\label{newopz}
\argmin_{\phi \in \mathbb{R}^n} \sum_{S \in 2^N} \left[\sum_{i\in S}\phi_i-\left\{v_{\tau}(S)-v_{\tau}(\emptyset)\right\}\right]^2\pi_{x_{\tau}}(S). 
\end{equation}

Note that (\ref{newopz}) is essentially same as the optimization problem for LIME in \cite{lime}.
Furthermore, \cite{lundberg2017} impose a local accuracy condition (or called efficiency condition) on this optimization problem: $f(x)=g(x)=\phi_0+\sum_{i=1}^n\phi_ix_i$ for $x$. If this is imposed on  (\ref{newopz})  and letting $\Psi_{\tau}^{AFA}$ be the solution of this problem, the problem becomes as follows: 

\begin{equation}\label{trueop}
\Psi_{\tau}^{AFA} = \argmin_{\phi \in \mathbb{R}^n with \sum_{i\in N}\phi_i = v_{\tau}(N)-v_{\tau}(\emptyset)} \sum_{S \in 2^N} \left[\sum_{i\in S}\phi_i-\left\{v_{\tau}(S)-v_{\tau}(\emptyset)\right\}\right]^2\pi_{x_{\tau}}(S). 
\end{equation}
 
For the following analysis, we derive analytical solutions to the minimization problems of (\ref{newopz}) and (\ref{trueop}) by imposing a symmetric condition (Subsections \ref{Kcondtns} to \ref{ASOP}). Then we particularly focus on the solution of (\ref{trueop}) to propose some AFAs alternative to SHAP and to compare them with SHAP in terms of the associated kernels $\pi_{x_{\tau}}(S)$ (Subsections \ref{SHAPKenl} to \ref{lastAFAAS}). Note that, for both the unconstrained minimization (\ref{newopz}) and the constrained minimization  (\ref{trueop}), it is obvious that scalar multiplication of the kernel does not alter the minimization result.

\section{Results}
\subsection{Symmetric Condition on Kernel}\label{Kcondtns}
Regarding the kernel $\pi_{x_{\tau}}(S)$ in (\ref{newopz}) and (\ref{trueop}), we impose the following symmetric condition: 
\begin{eqnarray}
\pi_{x_{\tau}}(S) = \pi_{x_{\tau}}(T) \qquad \Big(\forall S,T\in 2^N \ with \ |S|=|T|\Big) \label{sym} 
\end{eqnarray}
(\ref{sym}) states that, in terms of the number of features, when $S$ and $T$ are equidistant from $N$, the kernel must assign the same weight to $S$ and $T$. This can be considered a form of symmetry, which is a naturally acceptable condition.
As mentioned above, the proportional scaling of the kernel does not change the minimization result. This implies that it may be beneficial to have a normalization condition. We will discuss this point in the following subsection.

\subsection{Analytical solution to the optimization problem with no constraint}\label{withnoconstsub}
We first derive the solution to the optimization problem (\ref{newopz}) where the efficiency condition is not imposed. The only substantial difference from \cite{lime} is that we impose the symmetric condition of (\ref{sym}) on the kernel  $\pi_{x_{\tau}}$. 

The F.O.C. on $\phi_j$ is:  
\begin{equation}
\displaystyle\sum_{S\in 2^N: j\in S}2
\left(\sum_{i\in S}\phi_i-\{v_{\tau}(S)-v_{\tau}(\emptyset)\}\right)\pi_{x_{\tau}}(S) = 0. \label{eq:Foc}
\end{equation}
Therefore, for any $i,j\in N$ with $i\neq j$, the following holds:
\begin{eqnarray*}
& \ & \sum_{S\in 2^N: i\in S}
\left(\sum_{k\in S}\phi_k-\{v_{\tau}(S)-v_{\tau}(\emptyset)\}\right)\cdot \pi_{x_{\tau}}(S) =\sum_{S\in 2^N: j\in S}
\left(\sum_{k\in S}\phi_k-\{v_{\tau}(S)-v_{\tau}(\emptyset)\}\right)\cdot \pi_{x_{\tau}}(S) \\
& \Longleftrightarrow  & \sum_{S\subseteq N\setminus \{i,j\}}\left(\sum_{k\in S\cup \{i\}}\phi_k-\{v_{\tau}(S\cup \{i\})-v_{\tau}(\emptyset)\}\right)\cdot \pi_{x_{\tau}}(S\cup \{i\}) \\
& \ & \qquad \qquad \qquad \qquad \qquad \qquad \qquad =\sum_{S\subseteq N\setminus \{i,j\}}
\left(\sum_{k\in S\cup \{j\}}\phi_k-\{v_{\tau}(S\cup \{j\})-v_{\tau}(\emptyset)\}\right)\cdot \pi_{x_{\tau}}(S\cup \{j\}) \\
& \Longleftrightarrow  & \sum_{S\subseteq N\setminus \{i,j\}}
\bigg(\pi_{x_{\tau}}(S\cup \{i\})\cdot \phi_i-\pi_{x_{\tau}}(S\cup \{j\})\cdot \phi_j\bigg) \\
& \ & \qquad \qquad \qquad \qquad \qquad \qquad \qquad =\sum_{S\subseteq N\setminus \{i,j\}}
\bigg(\pi_{x_{\tau}}(S\cup \{i\})\cdot v_{\tau}(S\cup \{i\})-\pi_{x_{\tau}}(S\cup \{j\})\cdot v_{\tau}(S\cup \{j\})\bigg) \\
& \Longleftrightarrow  & \phi_i-\phi_j=\sum_{S\subseteq N\setminus \{i,j\}}
\bigg(\pi_{x_{\tau}}(S\cup \{i\})\cdot \{v_{\tau}(S\cup \{i\})-\pi_{x_{\tau}}(S\cup \{j\})\cdot \{v_{\tau}(S\cup \{j\})\bigg),
\end{eqnarray*}
which implies:
\begin{equation}\label{eajojo}
\displaystyle \phi_1-\sum_{S: 1\in S, S\neq N}\pi_{x_{\tau}}(S)\cdot v_{\tau}(S)
 = ... =
\displaystyle \phi_n-\sum_{S: n\in S, S\neq N}\pi_{x_{\tau}}(S)\cdot v_{\tau}(S).
\end{equation}
\ \\
Furthermore, from (\ref{eq:Foc}), it follows that: 
\begin{equation*}
\displaystyle\left(\sum_{S\in 2^N: j\in S}\pi_{x_{\tau}}(S)\right)\cdot \phi_{j}
+\sum_{i\in N: i\neq j}\left(\sum_{S\in 2^N: i,j\in S}\pi_{x_{\tau}}(S)\right)\cdot \phi_{i}
=\sum_{S\in 2^N: j\in S}\pi_{x_{\tau}}(S)\cdot \Big(v_{\tau}(S)-v_{\tau}(\emptyset)\Big).
\end{equation*}
Thefore, by summing both sides over all $j\in N$:
\begin{equation}
\displaystyle\left(\sum_{S\in 2^N: j\in S}\pi_{x_{\tau}}(S)
+(n-1)\cdot \sum_{S\in 2^N: i,j\in S}\pi_{x_{\tau}}(S)\right)\cdot \sum_{j\in N}\phi_{j}
=n\cdot \sum_{S\in 2^N: j\in S}\pi_{x_{\tau}}(S)\cdot \Big(v_{\tau}(S)-v_{\tau}(\emptyset)\Big).
\label{eq:sum}
\end{equation}
From (\ref{eq:Foc}) and (\ref{eq:sum}), $\phi =(\phi_1,...,\phi_n)$ is expressed as follows:
\begin{equation}\label{APfinalans}
\displaystyle{\phi_{j}= 
\sum_{S: j\in S\neq N} \pi_{x_{\tau}}(S)\cdot v_{\tau}(S)
+\frac{T-\sum_{i\in N}
\left\{\sum_{S: i\in S\neq N}\pi_{x_{\tau}}(S)\cdot v_{\tau}(S)\right\}}{n}}
\end{equation}
where 
\begin{equation}\label{whereT}
T=\displaystyle \frac{n\cdot \sum_{S\in 2^N: j\in S}\pi_{x_{\tau}}(S)\cdot \Big(v_{\tau}(S)-v_{\tau}(\emptyset)\Big)}{\sum_{S\in 2^N: j\in S}\pi_{x_{\tau}}(S)+(n-1)\cdot \sum_{S\in 2^N: i,j\in S}\pi_{x_{\tau}}(S).}
\end{equation}

The analytical solution consists of (\ref{APfinalans}) and (\ref{whereT}). Here it should be noted that, \cite{lime} considers a general case where the penalty term $\Omega(z)$ is non-zero and does not seek to derive an analytical solution of the minimization problem. Instead, it proposes an algorithm to find a solution approximately. By focusing on the zero penalty case and imposing the symmetric condition on the kernel, our analysis succeeds in deriving an analytical solution of this problem. It should also be noted that this solution is a generalization of the solution of the optimization problem with the efficiency condition, which we will examine in the next subsection.


\subsection{Analytical solution to the optimization problem with the efficiency constraint}\label{ASOP}
Next, we derive $\Psi_{\tau}^{AFA}$ in (\ref{trueop}) analytically.\footnote{
In the context of the cooperative game theory, \cite{ruiz1998} examined a similar but distinct minimization problem with the efficiency condition, and in solving this problem, it pointed out that the optimal solution to the problem is unchanged if the problem is simplified in a certain way. This simplified problem is identical to our minimization problem (\ref{trueop}).} The Lagrangian of (\ref{trueop}) is: 
\begin{equation*}
\mathcal{L}(\phi_1,...\phi_n, \lambda) = 
\sum_{S\in 2^N}\left[\sum_{i\in S}\phi_i-\{v_{\tau}(S)-v_{\tau}(\emptyset)\}\right]^2 \cdot \ \pi_{x_{\tau}}(S) -\lambda\left[\sum_{i\in N}\phi_i-v_{\tau}(N)+v_{\tau}(\emptyset)\right].
\end{equation*}
The F.O.C. on $\phi_j$ is:  
\begin{equation*}
\displaystyle\sum_{S\in 2^N: j\in S}2
\left(\sum_{i\in S}\phi_i-\{v_{\tau}(S)-v_{\tau}(\emptyset)\}\right)\cdot \ \pi_{x_{\tau}}(S) - \lambda = 0,
\end{equation*}
which implies (\ref{eajojo}) holds, as in Subsection \ref{withnoconstsub}.
Therefore, $\phi =(\phi_1,..., \phi_j,..., \phi_n)$ that satisfies (\ref{eajojo}) and $\sum_{j\in N}\phi_j=v_{\tau}(N)-v_{\tau}(\emptyset)$ is derived as:
\begin{equation}\label{unifi}
\displaystyle{\Psi_{\tau, j}^{AFA} =\phi_{j}= \sum_{S: j\in S} \pi_{x_{\tau}}(S)\cdot v_{\tau}(S)+\frac{v_{\tau}(N)-v_{\tau}(\emptyset)-\sum_{i\in N}\left\{\sum_{S: i\in S}\pi_{x_{\tau}}(S)\cdot v_{\tau}(S)\right\}}{n}}.
\end{equation}

Some remarks are made. First, the efficiency constraint $\sum_{i\in N}\phi_i = v_{\tau}(N)-v_{\tau}(\emptyset)$ is essentially identical to assuming $\pi_{x_{\tau}}(N) = \infty$, and if so, (\ref{whereT}) holds with $T = v_{\tau}(N)-v_{\tau}(\emptyset)$. Therefore, (\ref{APfinalans}) coincides with (\ref{unifi}), i.e., (\ref{APfinalans}) is a generalization of the solution (\ref{unifi}) for the optimization problem with the efficiency condition. Second, (\ref{unifi}) expresses the AFA, $\Psi_{\tau}^{AFA}$, as a function of the associated kernels $\pi_{x_{\tau}}(S)$. 
This enables us to construct an AFA from any kernels that satisfy the symmetry condition, which is definitely powerful, as we will see.
In the following sections, we examine several AFAs, some proposed by existing research, while others are newly proposed and generated by kernels having appropriate properties.

\subsection{SHAP}\label{SHAPKenl}
In \cite{lundberg2017}, the kernel of SHAP is specified as follows:
\begin{equation}\label{niseKSS}
\pi_{x_{\tau}}(S)=\displaystyle{\frac{n-1}{_{n}C_{|S|}\cdot |S|\cdot (n-|S|)}}.
\end{equation}
Instead, we prefer the following rescaled kernel that satisfies our standardization condition discussed in Section \ref{Kcondtns}.
\begin{equation}\label{KSS}
\pi_{x_{\tau}}(S)=\displaystyle{\frac{n}{_{n}C_{|S|}\cdot |S|\cdot (n-|S|).}}
\end{equation}
By substituting (\ref{KSS}) into (\ref{unifi}), we obtain the following: 
\begin{equation*}
\Psi_{\tau, j}^{SHAP} = \phi_j = \displaystyle\sum_{S\subseteq N\setminus j}
\frac{|S|!(n-|S|-1)!}{n!}\left(v_{\tau}(S\cup \{j\}) - v_{\tau}(S)\right). \label{SHAP}
\end{equation*}
That is, SHAP is derived as an AFA generated from the kernel expressed in (\ref{KSS}). Therefore, it may be more appropriate to consider (\ref{KSS}) rather than (\ref{niseKSS}) as the kernel for SHAP. Additionally, it should be noted that the kernel of (\ref{niseKSS}) or (\ref{KSS}) reaches its maximum if $|S|=0$ and $|S|=n$, and it has a concave shape regarding $|S|$, which is different from \cite{lime} where the weight assigned by the kernel increases as a perturbed sample gets closer to the instance being explained.\footnote{Note that the value of kernel at $S=\emptyset$ does not matter for the minimization problem as long as we adopt the convention that $0 \times \infty = 0$.} 

\subsection{ES and FESP in \cite{Condevaux2023}}
As alternative AFAs to SHAP, \cite{Condevaux2023} proposes ES (Equal Surplus) and FESP (Fair Efficient Symmetric Perturbation), based on the solution concepts in cooperative game theory. 

First, consider the following kernel: 
\begin{equation}\label{ESkernel}
\pi_{x_{\tau}}(S)=\left\{\begin{array}{ll}
1 & \mbox{if}\quad |S|=1 \\ 
0 & \mbox{if}\quad 2\le |S| \le n.\\\end{array}\right.
\end{equation}
Similarly to the previous case, by substituting (\ref{ESkernel}) into (\ref{unifi}), $\phi_i$ becomes as follows:
\begin{equation*}\label{ESformal}
\Psi_{\tau, j}^{ES} = \phi_j = \displaystyle v_{\tau}(\{j\})+\frac{v_{\tau}(N)-v_{\tau}(\emptyset)-\sum_{k\in N}v_{\tau}(\{k\})}{n}.
\end{equation*}
That is, $\phi_i$ coincides with ES. 

Next, suppose the following kernel:
\begin{equation}\label{FESPkernel}
\pi_{x_{\tau}}(S)=\left\{\begin{array}{ll}
w_{\tau} & \mbox{if}\quad |S|=1 \\ 
0 & \mbox{if}\quad 2\le |S| \le n-2 \\
1-w_{\tau} & \mbox{if}\quad  n-1 \le |S| \le n \ \end{array}\right.
\end{equation}
Then, (\ref{unifi}) follows that the associated solution of the minimization problem is FESP: 

\begin{equation*}
\Psi_{\tau, j}^{FESP} = \phi_j =  \displaystyle w_{\tau}\Big(v_{\tau}(\{j\})-v_{\tau}(\emptyset)\Big)+(1-w_{\tau})\Big(v_{\tau}(\emptyset)-v_{\tau}(N\backslash \{j\})\Big).
\end{equation*}

Note that the kernels of (\ref{ESkernel}) and (\ref{FESPkernel}) do not have the property that the weight of a perturbed sample increases as it gets closer to the instance of interest, i.e., as $|S|$ increases.
 
\subsection{AFA based on LS preucleolus}
Consider the following kernel: 
\begin{equation}\label{PreN}
\pi_{x_{\tau}}(S)=
\displaystyle{\frac{1}{2^{n-2}}} 
\end{equation}
Note that the shape of this kernel is not concave with respect to $|S|$, although it is still different from \cite{lime} in that the shape is flat. By substituting (\ref{PreN}) into (\ref{unifi}), the resulting $\phi_i$ is: 
 
\begin{equation*}
\Psi_{\tau, j}^{PNucl} = \phi_j = \displaystyle 
2\left(\frac{1}{2^{n-1}}\displaystyle\sum_{S: j\in S} v_{\tau}(S)\right)
+\frac{v_{\tau}(N)-v_{\tau}(\emptyset)-\sum_{i\in N}
\left\{2\left(\frac{1}{2^{n-1}}\sum_{S: i\in S}v_{\tau}(S)\right)\right\}}{n} \label{MNAP}
\end{equation*}

Intriguingly, this solution is identical to that in the following minimization problem in which a kernel does not appear, coinciding with the LS prenucleolus proposed by \cite{ruiz1996}:
\begin{equation*}\label{argmin}
\displaystyle\argmin_{\phi\in R^{n}: \sum_{i\in N}\phi_i=v_{\tau}(N)-v_{\tau}(\emptyset)}
\sum_{S\in 2^N\setminus \emptyset}\left[\sum_{i\in S}\phi_i-\{v_{\tau}(S)-v_{\tau}(\emptyset)\}\right]^2.
\end{equation*}

\subsection{AFA with a reasonable kernel (I)}
The next kernel we consider is as follows: 
\begin{equation}\label{sonono1}
\pi_{x_{\tau}}(S)=\displaystyle{\frac{|S|}{n\cdot 2^{n-3}}} 
\end{equation}
This kernel satisfies the conditions of  (\ref{sym}). Furthermore, this is increasing in $|S|$ and thus consistent with the condition on the kernel in \cite{lime}.
By substituting (\ref{sonono1}) into (\ref{unifi}), we have:
\begin{equation*}
\Psi_{\tau, j}^{LnK} =\displaystyle{\phi_{j}= 
\sum_{S: j\in S} \frac{|S|}{n\cdot 2^{n-3}}\cdot v_{\tau}(S)
+\frac{v_{\tau}(N)-v_{\tau}(\emptyset)-\sum_{i\in N}
\left\{\sum_{S: i\in S}\frac{|S|}{n\cdot 2^{n-3}}\cdot v_{\tau}(S)\right\}}{n}},
\end{equation*}
which is the first AFA we propose as an alternative to SHAP. The superscript $LnK$ stands for linealy increasing kernel. 

\subsection{AFA with a reasonable kernel (II)}
In \cite{lime},  the kernel associated with LIME is defined as follows:
\begin{equation*}
\pi_{x_{\tau}}(z)=\exp\left(\frac{-D(x,z)^2}{\sigma^2}\right)
\end{equation*}
where $D$ is a distance function with width $\sigma$. Recall that $x$ is the instance of interest and $z$ is a perturbed sample from $x$, and that, just for simplicity, these are assumed binary from the biginning. Therefore, following our notations and  the assumption (\ref{sym}), the distance function can be written by: 
\begin{equation*}
D(x,z)=\sqrt{\sum_{i\in S}0^2+\sum_{i\notin S}1^2}=\sqrt{n-|S|}.
\end{equation*}
Therefore, its associated kernel is $\pi_{x_{\tau}}(S)=\exp\left([-(n-|S|)]/\sigma^2\right)
=(e^{\frac{1}{\sigma ^2}})^{|S|}/(e^{\frac{1}{\sigma ^2}})^{n}$.
When normalizing this kernel, ensuring that the solution of the optimization problem (\ref{trueop})  in which the kernel is substituted remains unchanged, we obtain the following:
\begin{equation*}
\pi_{x_{\tau}}(S)=
\displaystyle{\frac{\left(e^{\frac{1}{\sigma ^2}}\right)^{|S|-1}}{\left(e^{\frac{1}{\sigma ^2}}+1\right)^{n-2}.}} 
\end{equation*}
Furthermore, by assuming $\sigma=\sqrt{1/\log 2}$, the following simplified LIME-type kernel is obtained:
\begin{equation}
\pi_{x_{\tau}}(S)=\displaystyle{\frac{2^{|S|-1}}{3^{n-2}.}} \label{shitatotsu}
\end{equation}
This kernel is increasing in $|S|$. More specifically, each time $|S|$ increases by $1$, the value of the kernel doubles.
Then, we get the following expression, which is our second proposed AFA alternative to SHAP.  
\begin{equation*}
\Psi_{\tau, j}^{ExK} = \displaystyle{\phi_{j}= \sum_{S: j\in S} \frac{2^{|S|-1}}{3^{n-2}}\cdot v_{\tau}(S) +\frac{v_{\tau}(N)-v_{\tau}(\emptyset)-\sum_{i\in N}
\left\{\sum_{S: i\in S}\frac{2^{|S|-1}}{3^{n-2}}\cdot v_{\tau}(S)\right\}}{n}}
\end{equation*}
The superscript $ExK$ stands for exponentially increasing kernel.

\subsection{AFA with a reasonable kernel (III)}\label{lastAFAAS}
The kernel of type (\ref{PreN}) and (\ref{sonono1}) correspond to the uniform kernel and the triangualr kernel, respectively. (\ref{shitatotsu}) can be regarded as a convex kernel function. Contrasting to thoese kernels, we lastly consider the following concave kernel function, corresponding to Epanechnikov or cosine kernel.  

\begin{equation}
\pi_x(S)=\displaystyle\frac{|S|(2n-|S|)}{(3n^2-n+2)\cdot 2^{n-4}.} \label{concavekernel}
\end{equation}
In this case, the solution of (\ref{unifi}) becomes as follows:

\begin{equation*}
\Psi_{\tau, j}^{Cncav} = \displaystyle{ \sum_{S: j\in S} \frac{|S|(2n-|S|)}{(3n^2-n+2)\cdot 2^{n-4}}\cdot v_{\tau}(S)+\frac{v_{\tau}(N)-v_{\tau}(\emptyset)-\sum_{i\in N}\left\{\sum_{S: i\in S}\frac{|S|(2n-|S|)}{(3n^2-n+2)\cdot 2^{n-4}}\cdot v_{\tau}(S)\right\}}{n}}.
\end{equation*}

If a kernel function is convex as (\ref{shitatotsu}), it implies that the weight associate with $z$ substantially drops for a small deviation from $x$ of the instance of interest. 
If a kernel function is concave as  (\ref{concavekernel}), it implies that the decline in the weight of $z$ is limited for the same deviation from $x$. 

\section{Conclusion}
In this study, we first derive an analytical and general expression of an AFA as a function of its associated kernel. Next, we compute several AFAs based on representations of several different specific kernels. Among the existing AFAs, we show that for SHAP, by slightly modifying the kernel into an appropriate form, the generated AFA coincides with SHAP. Additionally, for ES and FESP, we derive the representations of the corresponding kernels. The last four kernels and the AFAs generated from them are proposed for the first time in this study. $\Psi_{\tau}^{PNucl}$ has a kernel that is not concave and coincides with the notion of the LS prenucleolus in the cooperative game theory. $\Psi_{\tau}^{LnK}$, $\Psi_{\tau}^{ExK}$, and $\Psi_{\tau}^{Cncav}$ are generated from kernels that have desirable properties and consistent with the idea from \cite{lime} that the kernel assigns a large weight as a perturbed sample gets closer to the instance being explained.

The extent to which these AFAs show different decomposition patterns in experiments using actual data is an empirical question of great importance and one that should be addressed promptly. Another important theme is how the newly presented $\Psi_{\tau}^{LnK}$, $\Psi_{\tau}^{ExK}$, and $\Psi_{\tau}^{Cncav}$ in this study can be characterized from the perspective of cooperative game theory, for example, whether they can be axiomatized, is also worth investigating.

\renewcommand{\thesection}{A}
\section*{Appendix: Some properties of AFA in relation to prediction models}
In this appendix, we present some properties of AFA defined in (\ref{unifi}), especially those in relation to prediction models $f$. In Subsection \ref{coincidence}, we demonstrate that if the prediction model is additive with respect to features, the AFA represented by (\ref{unifi}) becomes identical regardless of the shape of the kernels. In Subsection \ref{OLSsbs}, we examine the linear regression model as a special case of the additive prediction model, and show that all AFAs represented by (\ref{unifi}) coincide with the AFA derived from the parameters of the linear regression model. The former suggests that the various AFAs in our analysis (shown in Subsections \ref{SHAPKenl} to \ref{lastAFAAS}) lead to different patterns of factor decomposition  only when the prediction model is nonadditive. The latter confirms the validity of employing the AFAs expressed as (\ref{unifi}) through their relationship with linear regression models.

\subsection{Coincidence of AFAs in additive prediction models}\label{coincidence}
\begin{chr}\label{general2}
Suppose the prediction model $f$ is additive with respect to $X_j$, i.e., 
\begin{equation}\label{additivepredmodel}
Y= f(X) = \sum_{j=1}^n f_j(X_j).
\end{equation}
Then, $\Psi_{\tau}^{AFA}$ in (\ref{unifi}) is identical regardless of the kernel representations, and satisfies $\Psi_{\tau, j}^{AFA} = v_{\tau}(N)-v_{\tau}(N\backslash \{j\})$ for all $j$. 
\end{chr}
Property \ref{general2} means that, when the prediction model is additive, all AFAs examined in our analysis coincide. In other words, when the prediction model is non-additive, applying different AFAs leads to different results.

In order to prove Property \ref{general2}, we first present the following Definition \ref{additivedef} and Lemma \ref{additivelem2}.
\begin{defi}\label{additivedef}
If a characteristic function form game $(N, v)$ satisfies the following condition, then $(N, v)$ is called additive.
\begin{equation}\label{additivity}
\forall S, T \ with \ S\cap T =\emptyset, \ v(S\cup T)-v(\emptyset) = \{v(S)-v(\emptyset)\}+\{v(T)-v(\emptyset)\}. 
\end{equation}
When $(N, v)$ is additive, we also say that $v$ is additive. Note that  (\ref{additivity}) is equivalent to the following:
\begin{equation*}
\exists a = (a_1,...,.a_n)\in R^n, \ \forall S\in 2^N, \ v(S)-v(\emptyset) = \sum_{j\in S}a_j.
\end{equation*}
\end{defi}

\begin{lemm}\label{additivelem2}
If the prediction model $f$ is additive with respect to $X_j$, then the characteristic function form game $(N, v_{\tau})$ defined in (\ref{vtau}) is additive. 
\end{lemm}
\textbf{Proof of Lemma \ref{additivelem2}} \ \ 
Because the prediction model is expressed as (\ref{additivepredmodel}),  from (\ref{vtau}), it follows that:  
\begin{equation*}
v_{\tau}(S) =E\left[f(x_{\tau, S}, X_{N\setminus S}) \right]=
\sum_{k: k\in S} f_k (x_{\tau, k}) + \sum_{l: l\not\in S} E\left[f_l(X_l)\right].
\end{equation*}
Letting $a_j = f_ j (x_{\tau,j})-E \left[f_j(X_j)\right]$, for any $S\in 2^N$, the following holds:  
\begin{equation}
v_{\tau}(S)-v_{\tau}(\emptyset) = 
\left(\sum_{k: k\in S} f_k (x_{\tau, k}) + \sum_{l: l\not\in S} E\left[f_l(X_l)\right]\right)
- \sum_{j=1}^n E\left[f_j(X_j)\right] = \sum_{j:j\in S} \left(f_ j (x_{\tau,j})-E \left[f_j(X_j)\right]\right) = \sum_{j\in S}a_j. 
\ \ \ \blacksquare \notag
\end{equation} 

\ \\
\textbf{Proof of Property \ref{general2}} \ \ 
Suppose the prediction model $f$ is additive with respect to $X_j$. From Lemma \ref{additivelem2}, $(N, v_{\tau})$ is additive. Therefore, from (\ref{unifi}), it follows that:
\begin{eqnarray*}
\Psi_{\tau, i}^{AFA}-\Psi_{\tau, j}^{AFA} & = & \sum_{S\subseteq N\setminus \{i,j\}}
\bigg(\pi_{x_{\tau}}(S\cup \{i\})\cdot v_{\tau}(S\cup \{i\})-\pi_{x_{\tau}}(S\cup \{j\})\cdot v_{\tau}(S\cup \{j\})\bigg) \\
& = & \sum_{S\subseteq N\setminus \{i,j\}}
\bigg(\pi_{x_{\tau}}(S\cup \{i\})\cdot \sum_{k\in S\cup \{i\}}a_k
-\pi_{x_{\tau}}(S\cup \{j\})\cdot  \sum_{k\in S\cup \{j\}}a_k\bigg) = a_i-a_j.
\end{eqnarray*}
Furthermore, since $\sum_{j\in N}\Psi_{\tau, j}^{AFA}= \sum_{j\in N}a_j$, it holds that $\Psi_{\tau, j}^{AFA}  = a_j$. Finally, since the following holds:  
\begin{equation}
v_{\tau}(N)-v_{\tau}(N\backslash \{j\}) = 
\sum_{k=1}^n  f_k (x_{\tau, k}) - \left(\sum_{k: k\neq j} f_k (x_{\tau, k}) + E\left[f_l(X_j)\right]\right) = f_ j (x_{\tau,j})-E \left[f_j(X_j)\right]=a_j, \notag
\end{equation} 
$\Psi_{\tau, j}^{AFA}  = v_{\tau}(N)-v_{\tau}(N\backslash \{j\})$. $\blacksquare$

\subsection{Coincidence of AFAs with parameters in linear regression models}\label{OLSsbs}
Next, suppose the prediction model $f$ is a linear regression model expressed as: 
\begin{equation}\label{OLScase}
Y= f(X) = \beta_0 +\sum_{j=1}^n \beta_jX_j.
\end{equation}

In this case, an AFA expression using the regression parameters is available. Namely, if the prediction model is expressed as (\ref{OLScase}), then
$v_{\tau}(S) =E\left[f(x_{\tau, S}, X_{N\setminus S}) \right]= \beta_0 +\sum_{k: k\in S} \beta_k x_{\tau, k} + E\left[\sum_{l: l\not\in S} \beta_lX_l\right].$
Therefore, the following holds:
\begin{equation*}
v_{\tau}(N)-v_{\tau}(\emptyset) =  \left(\beta_0 +\sum_{j=1}^n \beta_ jx_{\tau,j}\right) - \left(\beta_0 +\sum_{j=1}^n \beta_ jE\left[X_j\right]\right) = \sum_{j=1}^n \beta_ j \left(x_{\tau,j}-E\left[X_j\right]\right).
\end{equation*}
By using the regression parameters, if we define $\Psi_{\tau}^{LM} = (\Psi_{\tau,1},...,\Psi_{\tau,n})$ as  
\begin{equation}\label{afamlm}
\Psi_{\tau,j}^{LM} = \beta_ j \left(x_{\tau,j}-E\left[X_j\right]\right),
\end{equation}
then $\Psi_{\tau}^{LM}$ is AFA.  

\begin{chr}\label{general}
If the prediction model $f$ is a linear regression model, then $\Psi_{\tau}^{LM} =\Psi_{\tau}^{AFA}$ 
where $\Psi_{\tau}^{AFA}$ is expressed as (\ref{unifi}). 
\end{chr}
It is known that $\Psi_{\tau}^{LM}$ coincides with SHAP ($\Psi_{\tau}^{LM} = \Psi_{\tau}^{SHAP}$). Property \ref{general} shows that the same property holds for all other AFAs, including ones in our analysis, as far as it is expressed as (\ref{unifi}). This serves as one of the justifications for adopting the AFA expressed by equation (\ref{unifi}) to any prediction models, not limited to linear regression ones.

\ \\ 
\textbf{Proof of Property \ref{general}} \ \ \ Since a linear regression model is the additive model expressed as (\ref{additivepredmodel}), $\Psi_{\tau, j}^{AFA}  = a_j$ holds as shown in the proof of Property \ref{general2}. Furthermore, if the model is linear regression, it holds that   
$a_j = f_ j (x_{\tau,j})-E \left[f_j(X_j)\right] = \beta_ j \left(x_{\tau,j}-E\left[X_j\right]\right)$. From (\ref{afamlm}), $\Psi_{\tau,j}^{LM} = \Psi_{\tau, j}^{AFA}.$ $\blacksquare $

\subsection{Other properties}
Finally, we present two more properties of AFAs when the number of features is less than four, without proof.\footnote{For detailed proofs, please contact the authors.}

\begin{chr}\label{general3}
When $n\le 3$, $\Psi_{\tau}^{SHAP} = \Psi_{\tau}^{PNcul}$ holds for any prediction models $f$.
\end{chr}

\begin{chr}\label{general4}
When $n\le 2$, $\Psi_{\tau}^{AFA}$ in (\ref{unifi}) is identical regardless of the kernel representations for any prediction models $f$.
\end{chr}



\end{document}